\documentclass[journal]{IEEEtran}
\usepackage{graphicx}
\usepackage{cite}
\usepackage{picinpar}
\usepackage{amsmath}
\usepackage{url}
\usepackage{flushend}
\usepackage[latin1]{inputenc}
\usepackage{colortbl}
\usepackage{soul}
\usepackage{multirow}
\usepackage{pifont}
\usepackage{color}
\usepackage{alltt}
\usepackage[hidelinks]{hyperref}
\usepackage{enumerate}
\usepackage{siunitx}
\usepackage{breakurl}
\usepackage{pbox}
\usepackage{booktabs}
\usepackage{amsmath,amssymb}
\usepackage{algorithmicx,algorithm}
\usepackage{float}
\usepackage{hyperref}
\usepackage{threeparttable}

\hypersetup{hidelinks,
	colorlinks=true,
	allcolors=black,
	pdfstartview=Fit,
	breaklinks=true}

\begin{document}
\title{PCF-Grasp: Converting Point Completion to Geometry Feature to Enhance 6-DoF Grasp}

\author{
	\vskip 1em
	
	Yaofeng Cheng, \IEEEmembership{Student Member, IEEE}, Fusheng Zha, Wei Guo, Pengfei Wang, \\ Chao Zeng \IEEEmembership{Member, IEEE}, Lining Sun and Chenguang Yang, \IEEEmembership{Fellow, IEEE}

	\thanks{
		\textbf{This work has been submitted to the IEEE for possible publication. Copyright may be transferred without notice, after which this version may no longer be accessible.}
        
		This work was supported in part by the National Natural Science Foundation of China (U2013602, 51521003), National Key R\&D Program of China (2022YFB4601800), Self-Planned Task (2023FRFK01001) of State Key Laboratory of Robotics and System (HIT), National independent project of China under Grant (SKLR202301A12) and The Key Talent Project of Gansu Province.\textit{(corresponding authors: Lining Sun, Chenguang Yang.)}
		
		Yaofeng Cheng, Fusheng Zha, Wei Guo, Pengfei Wang, and Lining Sun are with the State Key Laboratory of Robotics and System at Harbin Institute of Technology. Harbin 150001, China. Fusheng Zha is also at Lanzhou University of Technology. (e-mail: chengyf@stu.hit.edu.cn; \{zhafusheng, wguo01, wangpengfei, lnsun\}@hit.edu.cn).
		
		Chao Zeng and Chenguang Yang are with the Department of Computer Science. University of Liverpool. L69 3BX Liverpool, U.K. (e-mail: chaozeng, cyang@ieee.org).
	}
}

 \maketitle
\begin{abstract}
The 6-Degree of Freedom (DoF) grasp method based on point clouds has shown significant potential in enabling robots to grasp target objects. 
However, most existing methods are based on the point clouds (2.5D points) generated from single-view depth images. 
These point clouds only have one surface side of the object providing incomplete geometry information, which mislead the grasping algorithm to judge the shape of the target object, resulting in low grasping accuracy. 
Humans can accurately grasp objects from a single view by leveraging their geometry experience to estimate object shapes. 
Inspired by humans, we propose a novel 6-DoF grasping framework that converts the point completion results as object shape features to train the 6-DoF grasp network. 
Here, point completion can generate approximate complete points from the 2.5D points similar to the human geometry experience, and converting it as shape features is the way to utilize it to improve grasp efficiency.
Furthermore, due to the gap between the network generation and actual execution, we integrate a score filter into our framework to select more executable grasp proposals for the real robot.
This enables our method to maintain a high grasp quality in any camera viewpoint. 
Extensive experiments demonstrate that utilizing complete point features enables the generation of significantly more accurate grasp proposals and the inclusion of a score filter greatly enhances the credibility of real-world robot grasping. 
Our method achieves a 17.8\% success rate higher than the state-of-the-art method in real-world experiments. Code and videos are available at \href{https://github.com/ChengYaofeng/PCF-Grasp}{https://github.com/ChengYaofeng/PCF-Grasp}.\end{abstract}


{}

\definecolor{limegreen}{rgb}{0.2, 0.8, 0.2}
\definecolor{forestgreen}{rgb}{0.13, 0.55, 0.13}
\definecolor{greenhtml}{rgb}{0.0, 0.5, 0.0}

\vspace{1em}
\noindent\textbf{Note:} This work has been submitted to the IEEE for possible publication. Copyright may be transferred without notice, after which this version may no longer be accessible.
\vspace{1em}

\section{Introduction}

\IEEEPARstart{T}{he} advancements in point cloud processing\cite{qi2017pointnet, qi2017pointnet++, wang2019dynamic} have provided a comprehensive understanding of point clouds for neural networks, particularly in terms of point geometry. This understanding has facilitated the emergence of 6-DoF grasp as a prominent grasping method, which is based on the point clouds, primarily due to its expansive grasp solution space. However, the points used for grasping are often generated by the single-view depth image called 2.5D points \cite{peng2021self, ten2017grasp, ma2023towards}, which are incomplete surface object points. Reflective areas sometimes exacerbate the incompleteness, resulting in many low-quality grasps (Fig.1, A). Additionally, due to limitations in the training process \cite{chen2023efficient, fang2020graspnet, wang2021graspness}, grasp network predictions often do not consider the motion planning required for a robot, restricting the robot to perform well only in a small workspace. These two crucial challenges still retard the robot 6-DoF grasping in real-world applications.

\begin{figure*}[!t]\centering
	\includegraphics[width=6.8in]{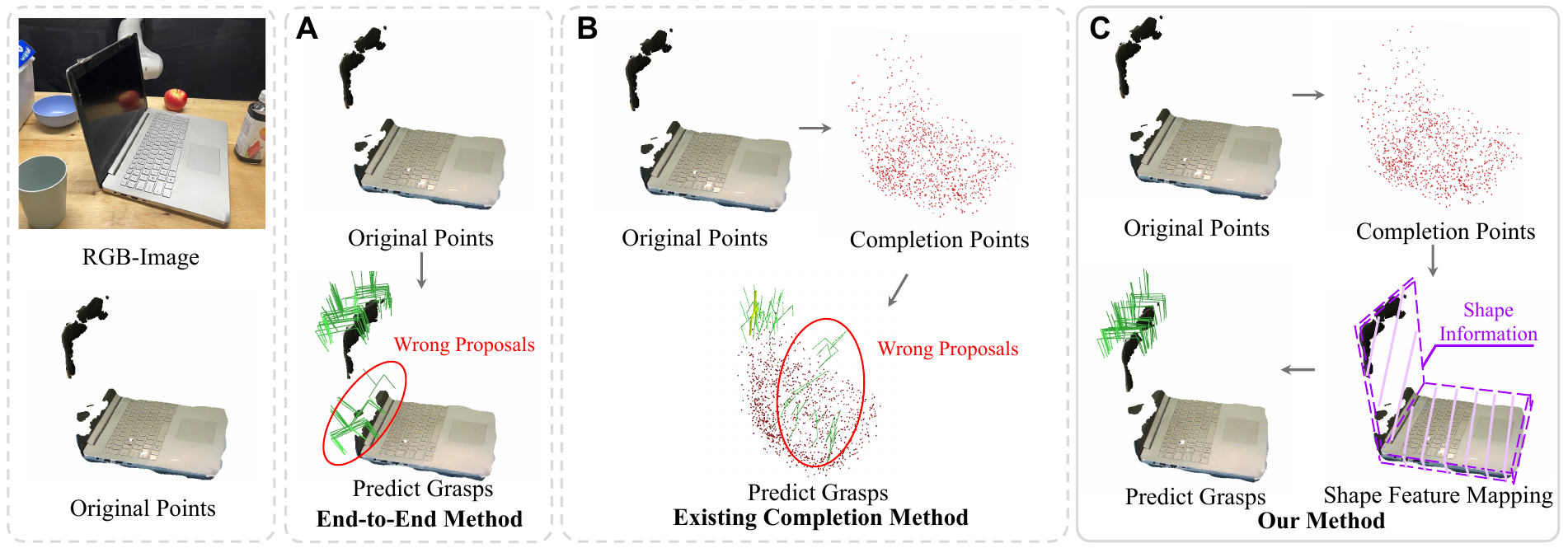}
	\caption{Difference between previous methods and our method. The computer is not trained in the point completion dataset leads to approximate completion results and its screen is reflective, where the depth camera loses its points. (A) The end-to-end method generates the wrong proposals at the bottom of the screen. (B) The existing completion method generates wrong proposals at the points where there is no corresponding object. (C) Our method remains robust by leveraging the point completion results as shape information, leading the network to generate precise grasps on the original points.}\label{fig_1}
\end{figure*}

Intuitively, capturing multi-view depth images or deploying multiple cameras around the object can overcome the issue of incomplete object points by capturing complete object points. Nevertheless, the placement of multiple cameras is cost-prohibitive, and it is time-consuming to mount the camera on a manipulator to rotate it around the object. In certain cases, terrain constraints prevent the collection of complete points. Hence, it becomes imperative to explore a solution to address the issue of incomplete points obtained from a single view. Several existing methods \cite{lundell2020beyond, mohammadi20233dsgrasp, keshari2023cograsp} have attempted to complete the entire object through point completion techniques, predicting the missing points from the visible ones, and facilitating the direct generation of grasps on the completed object points. However, the limitations of point completion techniques, specifically their dependency on optimal camera views and trained objects, can undermine precision and accuracy (Fig.1, B). This, in turn, significantly restricts the utility of point completion strategies in robotic grasping applications.

Inspired by \cite{munton2022see}, humans appear to utilize their geometric experience to approximate the shapes of objects by observing the visible parts. This ability enables them to execute reliable grasping processes. For example, when we observe a bottle, we intuitively perceive it as a cylindrical object in space and formulate grasping based on its visible parts. However, our observations suggest that humans primarily use this geometric experience for grasp judgment, evaluating whether an object's shape is suitable for grasping, while the decision of where to grasp relies directly on the visible areas of the object. 

Point completion techniques mirror the human use of geometric experience. Unlike previous methods that directly generate grasp points on completed results, we propose to encode them into shape features to enhance the grasping network, aligning more closely with human behavior. 
Specifically, we first employ a pre-trained point completion network to provide the approximate complete points for the target object. 
We then design a Point Completion to Feature Layer (PCF-Layer) to map the complete points as a shape feature within a hidden space and link this feature to the original incomplete points. 
In the final stage, we generate reliable grasp proposals by inputting both the original points and their associated shape features into a grasp network (Fig.1, C), ensuring the proposals are based on reliable original points.
Converting the point completion results into shape features provides additional whole object shape information from the pre-trained completion network. This not only enhances the grasp network's ability to recognize the target object --- enabling grasp predictions to account for the complete object --- but also alleviates the feature encoding burden on the point cloud encoder to some extent, allowing it to perform effectively with smaller datasets.

Moreover, the grasp pose is related to the points captured by the camera, establishing a relationship between grasps and camera views. Previous methods\cite{fang2020graspnet, mousavian20196, wang2019densefusion} usually grasp small objects, where the grasp directions vary in a small space. Some just place objects close to the robot or use an ideal camera view to keep the robot always in its workspace grasp in any direction. However, when the object is large, a little far or the camera view is not ideal for the robot, the various grasp directions, such as the grasping direction is opposite to the robot, will easily make the robot out of the workspace or collide with the target object. The grasp network, focused solely on end-effector poses, does not account for robot motion planning, thereby adversely affecting the success rate of grasp execution. Thus, based on the 6-DoF pose representation we use, we design a score filter using cosine similarity to filtrate the inconvenient grasp proposals for a real robot to reach. This significantly improves the grasping success rate in selecting an executable grasp for a robot from different camera viewpoints, preventing it from operating outside the workspace or colliding with objects.

To summarize, our key contributions are as follows:

\begin{enumerate}
	\item We propose a novel and, to the best of our knowledge, the first 6-DoF grasp framework that converts point completion results into features to train the 6-DoF grasp network. This approach enhances both grasp accuracy and generalization.

	\item We design a score filter to select more practical grasp poses from the grasp network output, bridging the gap between network predictions and real-world robot application, thereby improving the grasp success rate from any viewpoint.
	
	\item Real-world experiments demonstrate our framework achieves 89\% grasp success rate in real-world robot grasping, about 18$\%$ success rate higher than the state-of-the-art method.
\end{enumerate}

The rest of this paper is as follows. Section II reviews existing related works on robot grasping. Section III provides the details of our 6-DoF grasp framework. A series of experiments to evaluate the proposed method in Section IV. Conclusions and future work are drawn in Section V.


\begin{figure*}[!htb]\centering
	\includegraphics[width=7in]{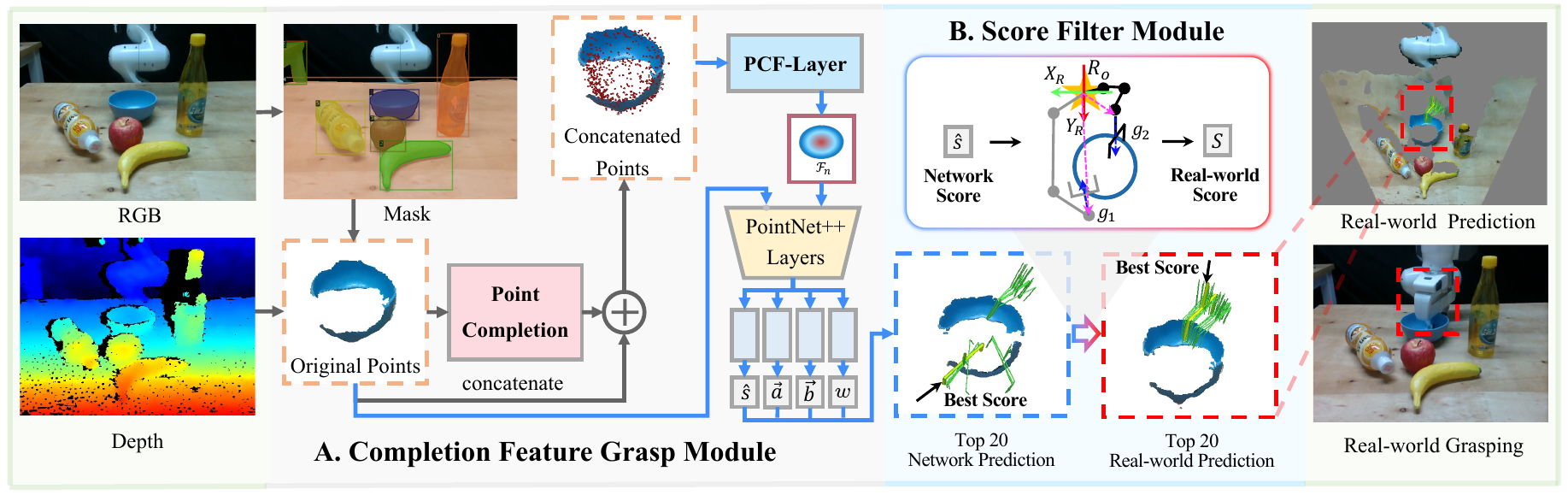}
	\caption{Pipeline of our whole proposed framework. (a) The Completion Feature Grasp Module is the grasp network prediction stage that predicts reasonable grasp proposals from the grasp network. Concatenating the completion points and the original points together and feeding them into the PCF-Layer to convert them to shape features for grasp training provide more object shape information for grasp prediction. (b) The score filter adjusts the grasp scores predicted by the network based on the positional relationship between the robot arm and the grasp pose, ensuring the selection of grasp poses that are more suitable for the robot arm. (c) The visualized grasps represent the top 20 scores selected from 1024 grasp candidates for the robot.}\label{fig_2}
\end{figure*}

\section{Related Works}
As a fundamental task in robotics, grasping has been studied for decades \cite{bicchi2000robotic, prattichizzo2016grasping, hang2014hierarchical}. In the following, we will focus on the aspects which are closely related to the proposed method.

\
\noindent {\bf{6-DoF Grasping:}} Classic data-driven methods \cite{lenz2015deep, wu2021real, shang2020deep} predict the rectangles using convolutional neural networks. Because these methods restrict the parallel gripper vertically to the camera view, the grasp results are not always as satisfying as people might expect. Hence, more and more works focus on the 6-DoF grasping research based on point clouds.

GPD\cite{ten2017grasp} formulates a new grasp representation for their CNN channels so that the CNN could directly predict the grasp pose through the RGB-D images. However, the prediction is restricted to the CNN detection results, which have less confidence in the box objects when the training objects mainly are spheres. Based on the PointNet, PointnetGPD\cite{liang2019pointnetgpd} proposes a network to sample the points within the gripper closing area for grasp representation. 6-DoF Graspnet\cite{mousavian20196} creates a dataset for some objects, which contains many successful grasp labels that are represented by $(R,T) \in SE(3)$ where $R \in SO(3)$ and $T\in \mathbb{R}^3$, and then uses an autoencoder to extract the grasp information to the network. The same as the \cite{mousavian20196}, CollisionNet \cite{murali20206} uses the autoencoder in their network, and they use the 3D bounding box to predict the object location first to check if there is any collision with the nearby objects. To avoid tedious preprocess, Contact Graspnet\cite{sundermeyer2021contact} directly trains the neural networks for grasp prediction by formulating the collision condition as the same problem as the grasp contact case. Graspnet-1Billion\cite{fang2020graspnet} offers a comprehensive 6-DoF grasping dataset across diverse objects, enabling the network to produce proposals based on geometric criteria. Graspness\cite{wang2021graspness} applies this concept to grasp in clutter scenarios. But all of these single-view methods are stuck into the incomplete point clouds captured by the camera and can not reach a high success rate.

\
\noindent {\bf{Completion for Robotic Grasping:}} As incomplete points are common in grasping situations, researchers recently proposed some methods to generate grasp poses by completing the object shape directly. \cite{varley2017shape} converts the 2.5D points into a mesh object, and then feeds the mesh data into a CNN Network to generate a complete mesh model. \cite{lundell2020beyond} completes the points into voxel mesh too. By placing the mesh objects into a physics simulator to get more viewpoints, they turn the FC-GQ-CNN\cite{satish2019policy}, a top-grasp planner, into a 6-DoF grasp pose planner. \cite{chen2022improving} uses a transformer encoder and a manifold decoder to turn the 2.5D points into completion points instead of the mesh models for 6-DoF grasp pose generation. There are still a lot of methods\cite{chavan2022simultaneous, gualtieri2021robotic, gao2021kpam, zhang2018learning} trying to complete the target object to generate high-quality grasp proposals. However, point completion methods are not ideal and can only get a coarse object shape that makes these works totally based on complete points for grasp limit to be used in very few scenarios.

\section{Method} 

Our goal is to generate reliable 6-DoF grasp proposals for objects present in a single view depth image, which is of a structured scene from any viewpoint. The overview of our framework is shown in Fig.2. We will introduce the details of this work as follows.

\subsection*{A. Method Overview}

By observing just one side of the target object, humans can generate a feasible grasp plan. We hypothesize that humans make an approximate shape judgment of the target object by viewing it from a single perspective. Consequently, finding a method to endow the network with the same capability would be an effective solution for improving the quality of grasp generation, particularly when dealing with incompletion points.

We proposed a novel 6-DoF grasp framework for robot grasping, which converts the point completion results into a shape feature to enhance the grasp network capability for incomplete points. As shown in Fig.2, our framework is comprised of two parts. A) The Completion Feature Grasp Module (CFG) is a 6-DoF grasp network that generates 6-DoF grasp proposals based on the Completion Point Feature ($\mathcal{F}$) outputted from the Point Completion to Feature (PCF-Layer) along with the incomplete original points. B) The Score Filter Module (ScF) serves as a crucial bridge between network proposals and actual robot execution. It operates by utilizing both the robot's position and the grasp proposals to filter and identify a set of executable grasps for the robot from CFG Module output.

\begin{figure}[!htb]\centering
	\includegraphics[width=2.5in]{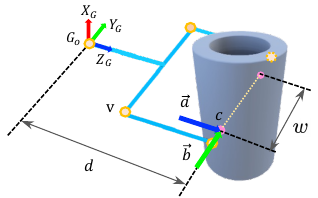}
	\caption{The grasp representation. $c$ depicts the grasp contact point. Vector $\mathbf{a}$ is the gripper approach, and $\mathbf{b}$ is the gripper grasp base. $w$ is the predicted grasp width, and $d$ is the distance from the base frame to the grasp baseline. The five orange points on the green gripper are the representation of the five gripper points $\mathbf v \in \mathbb R^{5 \times 3}$.}\label{fig_3}
\end{figure}

Concretely, we use the RGB image to obtain segmentation masks and then employ the target mask and the depth image to get the target's original points. These original points are input into a pre-trained point completion network to get completion points, which contain approximate spatial shape information. Subsequently, we concatenate these points with the original points to create a set of concatenated points that facilitate the mapping of shape features to the original points. By feeding the concatenated points into the PCF-Layer to acquire the shape feature $\mathcal{F}$, we then input the original points with their corresponding $\mathcal{F}$ into the PointNet++ encoder and a four-head MLP decoder to generate grasp proposals and scores. Following the output of the CFG's network prediction, the ScF identifies the most executable grasp proposals by considering both the robot's position and the grasp proposals.

\subsection*{B. Grasp Representation}

Efficient grasp representation makes it easier for learning-based methods, because of the difficulty of direct regression in high-dimensional space. Here, we follow the representation in \cite{sundermeyer2021contact} for parallel gripper grasp pose representation. It directly maps the grasps $g$ to the points $c \in {\mathbb{R}^3}$ that the gripper contacts the objects. Based on that, the grasp can be regarded as 3-DoF rotation $R_g \in \mathbb{R}^{3 \times 3}$ and grasp width $w \in \mathbb {R}$ of the gripper. And the 6-DoF grasp pose $g \in G$ can be defined by $(R_g, t_g) \in SE(3)$ and grasp width $w \in \mathbb R$ as

\begin{align}
t_{g}=\mathbf c\  +\  \frac{w}{2} \mathbf b\  +d \mathbf a
\end{align}
	
\begin{align}
R_g = \left[ \begin{matrix}\mid &\mid &\mid \\ \mathbf b& \mathbf a\times \mathbf b& \mathbf a\\ \mid &\mid &\mid \end{matrix} \right]  
\end{align}

where $\mathbf{a} \in \mathbb{R}^3$, $||\mathbf{a}|| = 1$ is the grasp approach vector, $\mathbf {b} \in \mathbb{R}^3$, $||\mathbf {b}|| = 1$ is the grasp baseline vector, $\mathbf c \in {\mathbb{R}^3}$ is the contact point between the gripper and the mesh, and $d\in \mathbb{R}$ is the constant distance from the gripper baseline to the gripper base. The grasp representation is given in Fig.3.

\begin{figure*}[!htb]\centering
	\includegraphics[width=6.6in]{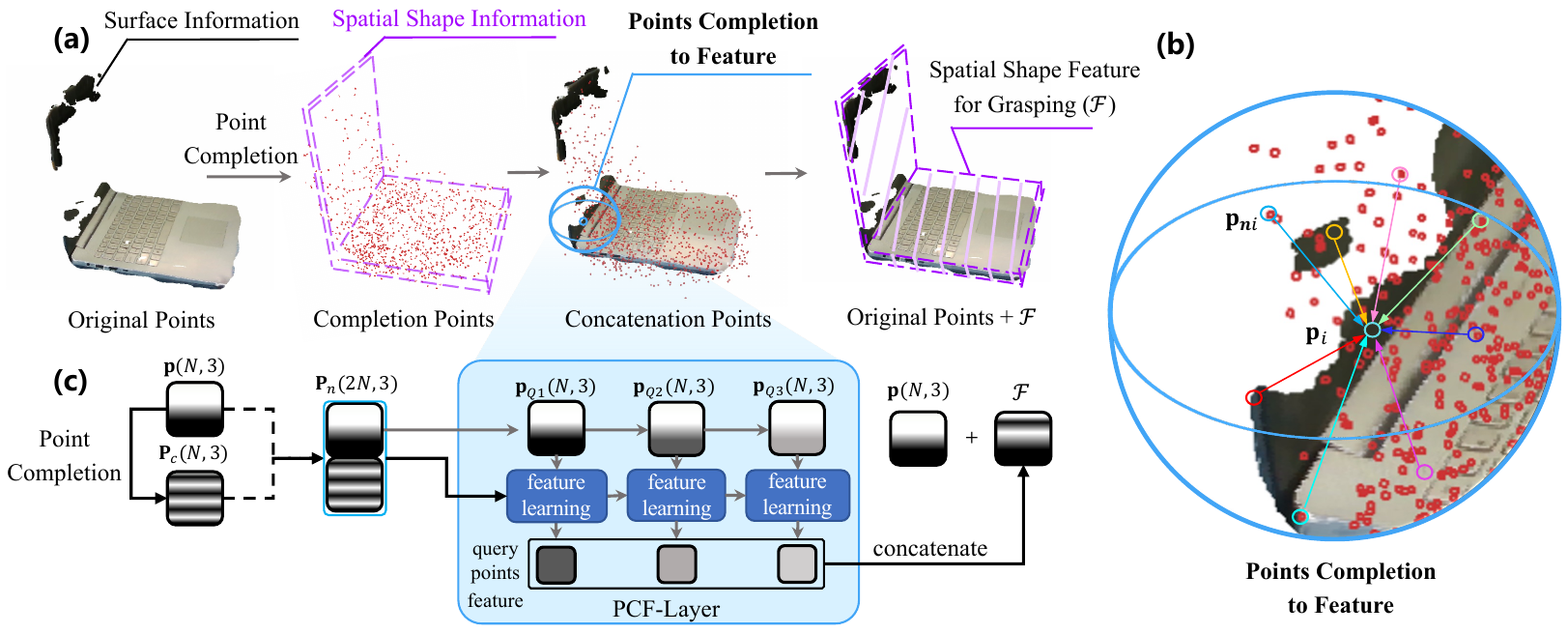}
	\caption{PCF-Layer mechanism. (a) In contrast to the original points, which only represent surface information, the completion points contain the object's approximate spatial shape information. Thus, the PCF-Layer provides the spatial shape features necessary for grasping. Notably, the purple shape outline for shape information and features is hand-crafted for easier understanding. (b) The PCF-Layer maps the concatenated points to the original points. (c) The PCF-Layer uses points feature learning blocks (drawing adapted from \cite{qi2017pointnet++, qi2017pointnet}) to extract the object shape information for grasp training.}\label{fig_4}
\end{figure*}

\begin{figure}[!htb]\centering
	\includegraphics[width=3.5in]{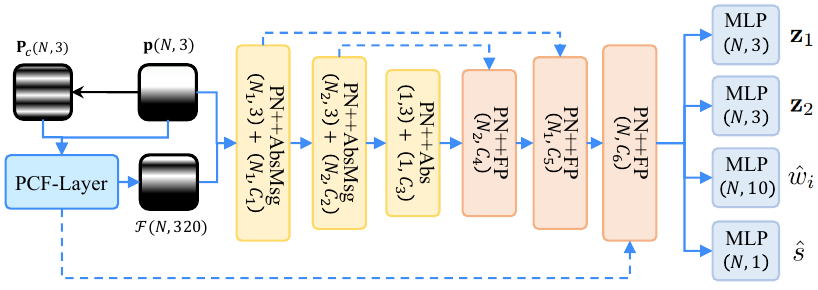}
	\caption{The grasp network. After the PCF-Layer outputs the $\mathcal{F}$, both it and the original points are fed into the point encoder to learn the complete points feature. Then four heads MLP decoders are used to generate grasp elements $\mathbf {z}_1$ and $\mathbf {z}_2$, grasp widths $\hat {w}_i$ and grasp scores $\hat {s}$.}\label{fig_5}
\end{figure}

\subsection*{C. Completion Feature Grasp Module}

Point completion can only approximate the shape of an object with coarse accuracy when the object is absent from the training dataset or the camera viewpoint is not ideal. Relying solely on point completion results to address the issue of low-quality grasp generation due to incomplete points restricts the applicability to grasp objects that are already presented in the point completion training data. Here, we convert the coarse completion points into shape features within a hidden space. This additional feature enables the grasp network to learn more effective grasping strategies by considering the entire object. Moreover, because the completion points are not absolutely precise, we generate reliable grasp proposals based on the original points.

\subsubsection{Point Completion to Feature}

After cropping the target original points, we select a number of $n=1024$ points $\mathbf p \in \mathbb{R}^{1024\times 3}$ with Farthest Points Sampling (FPS) as input for point completion and predict the complete object's coarse points $\mathbf P_c \in \mathbb{R}^{1024\times 3}$. Unlike the original points, which only capture surface details, the point completion results contain the approximate spatial shape of the object, providing more accurate shape information for grasping. Inspired by the point geometry learning method of PointNet++ \cite{qi2017pointnet++}, we design a PCF-Layer to extract the shape information for the grasp network. Although the point completion is trained based on original points, it differs from directly feeding the original points into the grasp network. This pre-trained point completion network learns point shape information through an additional convergence. With this extra shape information, the grasp network can develop better grasp generation capabilities. The structure of our PCF-Layer is shown in Fig.4. 

We concatenate the original points $\mathbf p$ and the completion points $\mathbf P_c$ to form the concatenated points $\mathbf P_{n} \in \mathbb{R}^{2048\times 3}$ representing the whole object points. We then use our PCF-Layer to map the concatenation points $\mathbf P_{n}$ as a feature to the original points $\mathbf p$. Specifically, as the points shape learning methods do \cite{qi2017pointnet, qi2017pointnet++}, we gather a number of $i=[64, 64, 128]$ closer points of $\mathbf {P_n}_i$ that $\in \mathbf P_{n}$ into three query ball of each original point$\mathbf p \in \mathbf p$ to groups $\mathbf p_{Q1}(N,3)$, $\mathbf p_{Q2}(N,3)$ and $\mathbf p_{Q3}(N,3)$. This is used to extract the shape of the concatenated points to provide more object shape information to the grasp network. Because the gripper width in our experiment is 0.08$m$, to match the grasp geometry, we set the query ball with radius $r=[0.04, 0.08, 0.16]m$. Following are 3 similar \cite{qi2017pointnet++} feature learning blocks, each block with $[32, 32, 64]$, $[64, 64, 128]$, $[64, 64, 128]$ MLP layers to learn the query ball points feature. After that, concatenate all the query ball points feature as $\mathcal{F} \in \mathbb{R}^{1024 \times 320}$. The points feature $\mathcal{F}$ contains the complete object points shape feature to lead to reasonable grasp training.

\subsubsection{Grasp Network}

With the complete object points feature $\mathcal{F}$ and the original points $\mathbf p$ as input, we follow \cite{sundermeyer2021contact} to train grasp prediction. The grasp network structure is shown in Fig.5. A PointNet++ segmentation structure layers to learn the points feature, then four heads to respectively predict the grasp confidence score $\hat s \in \mathbb{R}$, gripper width $\hat{w}_i \in [0, w_{max}]$ and two direction elements $\mathbf {z}_1 \in \mathbb {R}^3$, $\mathbf {z}_2 \in \mathbb {R}^3$. The input points for grasp prediction are the same $n=1024$ points $\mathbf p$ for points completion, which will be used as the actual grasp contact points for a robot.  As the\cite{zhou2019continuity} proposed, orthonormalization further facilitates the regression of 3D rotations. Thus, $\mathbf {z}_1$ and $\mathbf {z}_2$ represent the gripper approach $\hat {\mathbf a}$ and grasp baseline $\hat {\mathbf b}$ as Eq.(3), similar to \cite{sundermeyer2021contact}.

\begin{align}
	\mathbf {\hat{b}}=\frac{\mathbf z_1}{||\mathbf z_1||}\  \  \  \  \  \  \  \  \  \mathbf {\hat{a}}=\frac{\mathbf z_2 - \langle \mathbf {\hat{b}},\mathbf z_2 \rangle \mathbf {\hat{b}}}{||\mathbf z_2||}
\end{align}

With five gripper points $\mathbf v \in \mathbb R^{5 \times 3}$ shown in Fig.3, the predict grasp pose $v_i^{pred}$ and the ground truth grasp pose ${v}_i^{gt}$ can be written as

\begin{align}
	\mathbf {v}_i^{gt} = \mathbf {v}R_{g,i}^T + \mathbf t_{g,i}\  \  \  \  \  \  \  \  \  \mathbf v_i^{pred} = \mathbf {v}\hat R_{g,i}^T  + \hat{\mathbf {t}}_{g,i}
\end{align}

where $\hat R_{g,i}$ and $\hat t_{g,i}$ are the prediction rotation and transformation obtained by Eq.(1) and Eq.(2), $R_{g,i}$ and $t_{g,i}$ are the corresponding ground truth labels.

We define the grasp pose loss $\mathcal {L}_{add}$ as the distance between the ground truth gripper points $\mathbf {v}_i^{gt}$ and the prediction gripper points $\mathbf v_i^{pred}$. What's more, the symmetry of the gripper is considered. We use the closest ground truth grasp points' score $\hat {s}_i$ to weight the grasp distance loss $\mathcal L_{add-s}$. The loss function becomes:

\begin{align}
	\mathcal L_{add-s} = \frac{1}{n^+}\sum^{n^+}_i 
	\hat {s}_i \min_u ||\mathbf v_i^{pred}-\mathbf {v}_i^{gt}||_2
\end{align}

The score loss uses binary cross entropy as $\mathcal L_{bce}$ and the grasp width $\hat{w}_i$ is split into 10 constant widths $\hat {o} \in \mathbb{R}^{10}$ that would be considered as a classification problem with multi-label binary cross entropy loss as $\mathcal L_{width}$. The total loss is $\mathcal L_{total} = \alpha \mathcal L_{bce} + \beta \mathcal L_{add-s} + \gamma \mathcal L_{width}$ with $\alpha =1$, $\beta =10$, $\gamma=1$ same as \cite{sundermeyer2021contact}.

\begin{figure}[!htb]\centering
	\includegraphics[width=3.5in]{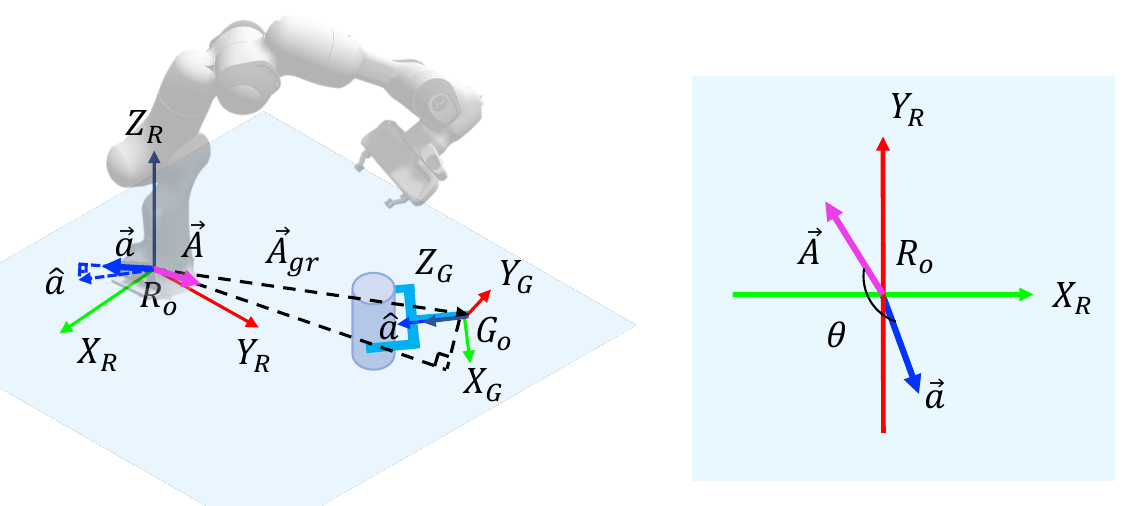}
	\caption{The score filter. $\mathbf {A}_{gr}$ is the vector from robots coordinate origin $R_o$ to the predict grasp $\hat g_i$ coordinate origin $G_o$, $\hat {\mathbf a}$ is the predicted gripper approach vector. The vector $\mathbf{A}$ and $\mathbf{a}$ on the right side is the unit vector of the projection of $\mathbf {A}_{gr}$ and $\hat {\mathbf a}$ on the plane $X_RR_oY_R$. $\theta$ is the angle between $\mathbf {A}$ and $\mathbf {a}$.}\label{fig_6}
\end{figure}

\subsection*{D. Score Filter Module}

The 6-DoF grasp offers flexibility for robot grasping due to its unrestricted directions. Nevertheless, supervised grasp training disregards the practical constraints of the robot, resulting in inconvenient grasp poses for grasping, such as out of workspace or collision with target object. In our approach, we select the grasp proposal with the highest confidence from the network outputs for execution, though this does not imply that alternative poses are inadequate. Many poses closely matching the top score are also suitable for robot grasping. For instance, our grasp network reliably outputs the top 20 grasp scores for successfully grasping objects. However, certain grasps may pose execution challenges, particularly if the grasp approaches are opposite to the robot, requiring extra joint manipulation and possibly exceeding the robot's workspace. It will fail when the best score is the inconvenient grasp pose. Therefore, it is necessary to filtrate those poses that are inconvenient for the robot to execute.

Based on the 6-DoF representation we utilize, we propose an ScF to filtrate the proposals generated by the grasp network which are not executable for robots. As shown in Fig.6, we consider the vector from the origin of the robot base coordinate $R_o$ to the origin of the predicted grasp pose $\hat {g}_i \in \hat G$ coordinate $G_o$ as $\mathbf{A}_{gr}$. Then we project $\mathbf{A}_{gr}$ and the grasp approach $\hat {\mathbf{a}}$ onto the robot base $X_RR_oY_R$ plane and convert them into unit vector $\mathbf{A}$ and $\mathbf{a}$.

\begin{align}
	\mathbf{A} = \frac{\mathbf{A}_{gr} - \mathbf{A}_{gr} \cdot Z_R}{||\mathbf{A}_{gr}-\mathbf{A}_{gr}\cdot Z_R||}\  \  \  \  \  \  \mathbf{a} = \frac{ R_{cr}(\mathbf{a} - \mathbf{a} \cdot Z_R)}{||R_{cr}(\mathbf{a} - \mathbf{a} \cdot Z_R)||}
\end{align}

where $\mathbf Z_R$ is the z-axis of the robot coordinate and $R_{cr}$ is the rotation of the camera relative to the robot.

To get reliable grasps, we consider the angle between the gripper approach direction $\mathbf{a}$ and the robot towards the gripper direction $\mathbf{A}$ try to be less than $90^{\circ }$. Therefore, we use Cosine Similarity to get a direction score $s_d$ between $\mathbf{a}$ and $\mathbf{A}$. Then, we use the sigmoid function to resize the direction score to $s_d \in [0,1]$.

\begin{align}
	s_d=sigmoid(\mathbf{A} \cdot \mathbf{a})
\end{align}

\begin{align}
	S=s_d \times \hat s   
\end{align}

At last, we weight the direction score to the score $\hat s$ generated by the network for the last grasp selection score $S$ and chose the highest score $S_{max}$ in $S$ as the grasp trial in real-world grasping. 

\subsection*{E. Implementation Details}

{\bf{Data Generation:}}
We place one object with dense grasp annotations from the ACRONYM dataset\cite{eppner2021acronym} with 20 categories of objects from the Shapenet dataset \cite{chang2015shapenet} at random stable poses in each scene. Depth images are rendered from the virtual camera from 48 randomly distributed viewpoints for generating the single-view point clouds that are used as the input points for both point completion and 6-DoF grasp. We totally generated 985 scenes. The completed object point clouds are created by sampling points uniformly on the mesh surfaces. We select 1024 points using FPS for coarse prediction.

{\bf{Point Completion Details:}}
We follow PCN\cite{yuan2018pcn} to finish the point completion and only train it with coarse prediction with 1024 points. We utilize AdamW optimizer\cite{loshchilov2018fixing} to train the network with an initial learning rate of 0.0001 and weight decay of 0.0005. We randomly choose 200 scenes from the above dataset for testing and the rest is used for training. The point completion network will be trained first with 300 epochs. Then the params will be fixed and used to provide complete points for the PCF-Layer.

{\bf{6-DoF Grasp Details:}}
After predicting grasps pose for each point of $\mathbf p$, to attach $\mathbf p$ to the annotation points, we consider the points as positive contacts if there exist label points in a $2mm$ radius and associate the select points belonging to the closest mesh contact to them. Because the datasets' successful labels always exist $\sim 100$ in each scene, we only select the top-k $k=108$ points with the largest errors as the grasp score head's backpropagation points.

We use AdamW optimizer with an initial learning rate of 0.0001 and a step-wise decay to 0.0005. In order to balance the Panda Emika gripper's grasp width and the whole object points characteristic, two PointNet++ layers are separately sampled $[64, 64, 128]$ points with query ball radius $[0.04, 0.08, 0.16]m$ and $[0.08, 0.16, 0.32]m$. The 6-DoF grasp network will be trained with 100 epochs.

\section{Experiment and Data Analysis}

In this section, we introduce the real robot experiment setting. Then, we present and analyze the results of real robot grasping experiments compared to state-of-the-art methods. Finally, extensive ablation studies are used to further evaluate each module in our proposed framework.

\subsection*{A. Experiment Setup}

We have evaluated our method with a 7-DoF Franka Panda robot where we pick both known category objects and unknown category objects in both ideal single-object scenes and cluttered scenes. Notably, the \textit{known \& unknown category} is different from the \textit{known \& unknown objects}. The category here means the category objects are trained \& not trained in the grasp dataset. The \textit{known \& unknown objects} means the objects are trained \& not trained in the point completion dataset. Therefore, a method that can grasp unknown categories must be able to grasp unknown objects. All objects in our experiments are unknown objects because we pick objects from daily life and not correspond to the dataset.

To ensure the realism of our grasping scenarios, we select 10 categories of objects with all kinds of materials as Fig.7 shows. In our experiments, we use bowl, apple, banana, bear and mouse as the unknown categories objects, and all kinds of bottles, laptop, wineglass, book and box as the known categories objects. Meanwhile, different from the traditional methods' objects, we chose all kinds of bottles and laptops that have extremely light or heavy weights, which would be pushed away or can't grasp up if the grasp pose is wretched. What's more, they have reflected materials that are sometimes difficult for the camera to capture.

\begin{figure}[!t]
\centering
\includegraphics[width=2.5in]{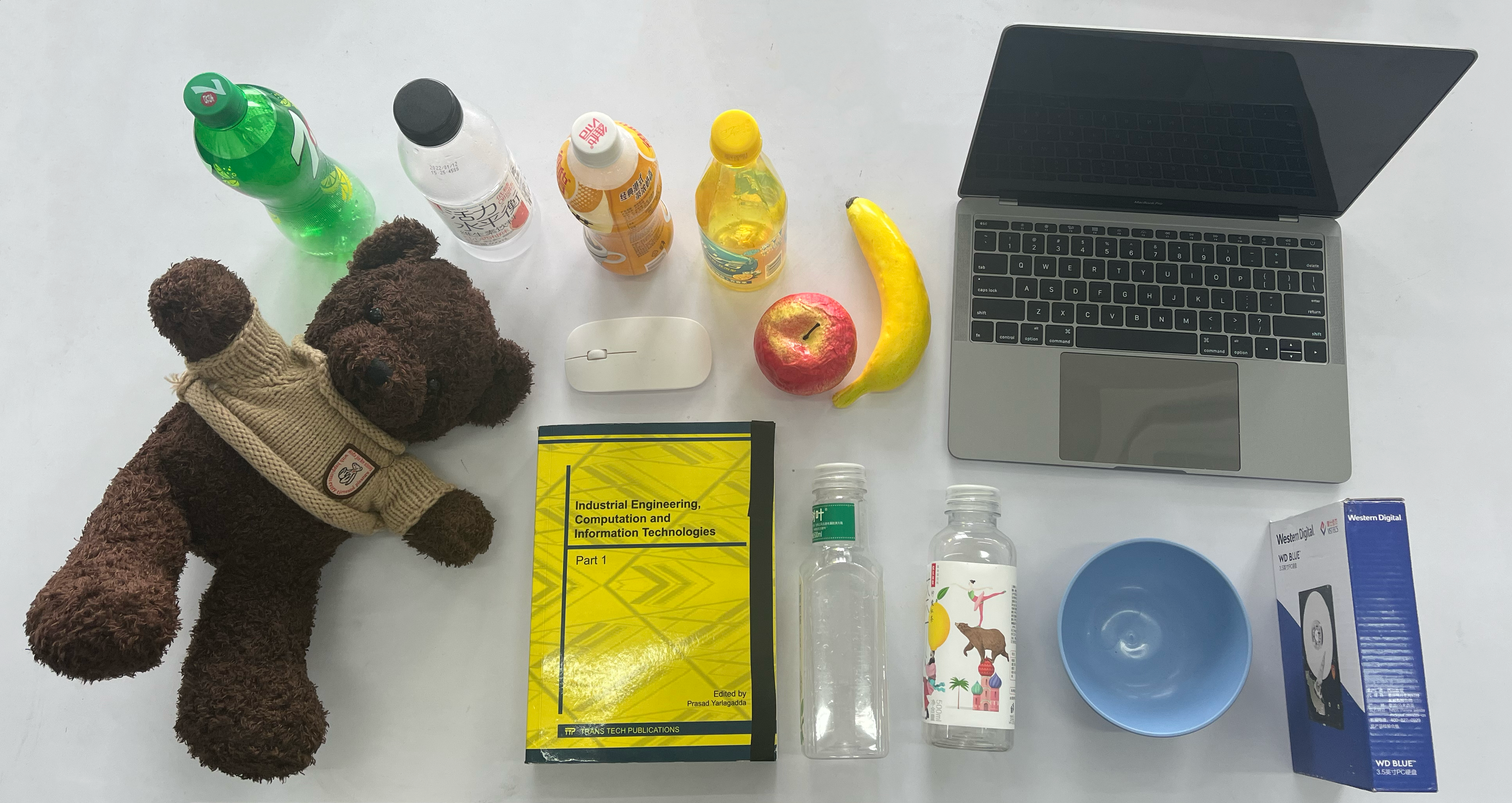}
\caption{Robotic Experiment Objects.}\label{fig_7}
\end{figure}

We use an Intel Realsense D435 camera mounted on a tripod to catch RGB images for segmentation mask using Detectron2 \cite{wu2019detectron2}, and depth images with the mask for generating original object points. In our experiments, our camera is opposite to the Franka robot (opposite side). Furthermore, to show the robustness of our algorithm, we also place the camera close to the robot (robot side), and on the right side of the robot (right side). The camera placement is shown in Fig.8. Each time we choose the highest score grasp pose to grasp the object.

We train and infer our model on a server computer (SC) configured as follows: Intel Core i9-10900X CPU, 3.7 GHz, Tesla V100 GPU 32 GB and 256 GB of RAM. Our robot grasp experiments are executed on a robot computer (RC) configured as follows: Intel Core i5-10400F CPU, 2.9 GHz, Ubuntu 18.04 RT kernel and 16 GB of RAM.

To further evaluate the importance of different components of our method, we define ablation study by comparing our method with follows:

\begin{itemize}
	\item Baseline: ours without point completion to feature and without ScF. The PCF-Layer is replaced by a PointNet++ layer. 
	\item Ours w/o PCF: ours without point completion to feature. To generate the grasp proposal based on the CP and our PCF-Layer is replaced by a PointNet++ layer.
	\item Ours w/o ScF: ours without ScF. The robot chooses the highest score from the CFG output.
	\item + ScF: use our ScF on other methods.
	\item Ours b/o completion: our method grasp is directly based on the PCN completion points.
	\item PC r/b FDNet: the point completion method replaced by FoldingNet \cite{yang2018foldingnet}
	\item PC r/b GRNet: the point completion method replaced by GRNet\cite{xie2020grnet}
	\item PC r/b PointR: the point completion method replaced by PointR\cite{yu2021pointr}
	\item PC r/b Disp3D: the point completion method replaced by Disp3D\cite{wang2022learning}
\end{itemize}

\subsection*{B. 6-DoF Grasp Experiments}

To verify the effectiveness of our proposed method, we contrast our method with some famous 6-DoF grasp methods: 6-DoF Graspnet\cite{mousavian20196}, PointNetGPD\cite{liang2019pointnetgpd}, Ma \textit{et al.}\cite{ma2023towards}, GraspNet-1Billion\cite{fang2020graspnet}, Graspness\cite{wang2021graspness} and Contact Graspnet\cite{sundermeyer2021contact}. Additionally, we also compare the completion-based grasp method: Beyond Top-Grasps \cite{lundell2019robust}, Trans SC \cite{chen2022improving}, 3DSGrasp \cite{mohammadi20233dsgrasp}. All the contrast experiments are implemented through the guidance of other papers' GitHub open-source code. After training, these methods generate grasp proposals when the point geometry is satisfying, even though the object category is not in the grasp dataset. Here, we use * to indicate the unknown category object in the corresponding grasp method in TABLE I.

\begin{figure}[!htb]\centering
	\includegraphics[width=2.5in]{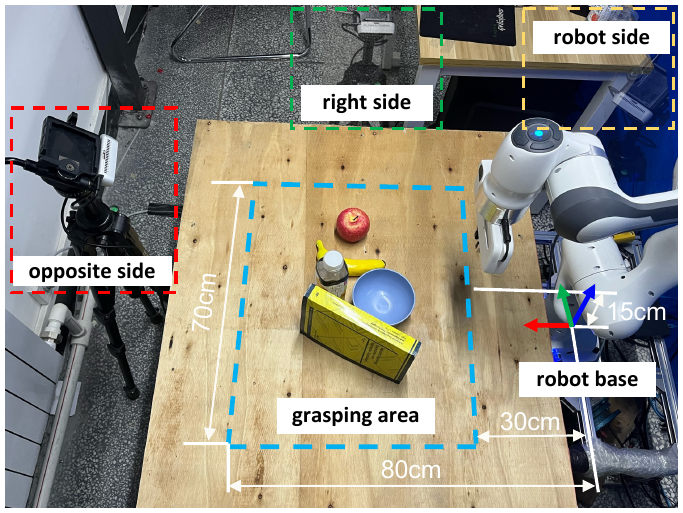}
	\caption{Different Camera viewpoints. Three camera viewpoints are used to demonstrate the generalization of our method for different camera viewpoints.}\label{fig_8}
\end{figure}

\
\subsubsection{Objects Placed in Isolation} In this experiment, all objects were placed in isolation, with the camera positioned opposite the robot. Each object was grasped only once for performance evaluation. Table I shows our real-world grasp success rates on the Franka robot. We have tried 50 times grasp for each object except for five bottles with different shapes, and each bottle has a total of 10 trials.

Our method utilizing the points completion feature has about 19.4$\%$ or more promotion of other end-to-end methods because of the more complete comprehension of the objects based on our point completion feature. Especially for laptops and bottles with reflective materials. Fig.9 (a) visualizes some sample predictions made by the state-of-the-art 6-DoF grasp method Contact Graspnet \cite{sundermeyer2021contact} and our whole methods. As we can see, Contact Graspnet failed to predict the grasp of the bottles in the rightmost column of (a) due to the reflective materials. In contrast, our method remains robust, which is because Contact Graspnet can only predict the grasp based on the visible points and ignore the complete object. But ours can take good advantage of the whole object geometry.

Compared to the completion-based method, our method has about 17.8$\%$ or more than other methods. As Beyond Top-Grasp is a 4-DoF grasp method with mesh completion, it cannot take full advantage of the completion results. Trans SC and 3DSGrasp achieve remarkable improvement on the trained objects in the point completion dataset, like banana, bowl, box, and book. But the untrained objects are not ideal, even worse. For example, with a larger geometric object like a laptop, the completion point cloud may be scattered between the keyboard and the screen, leading the grasping network to mistakenly interpret that part of the geometry as valid, resulting in a failed grasp. However, our method converts the point completion as a geometry feature to enhance the grasp prediction and the grasp proposals are based on the points captured by the camera. This not only avoids unreliable prediction points but utilizes the complete object geometry information from the point completion.

Another challenging case that can be seen in Fig.9 (a) and (b) is the direction of the highest score grasp (red grasp in the picture) in our method is always along the direction of the robot itself (location of the red pentagram), which avoids inconvenient grasp proposals for the robot to improve the grasp success rate. Further statistical analysis of object classes and sizes is provided in the supplementary materials.

\begin{figure*}[!htb]\centering
	\includegraphics[width=6.8in]{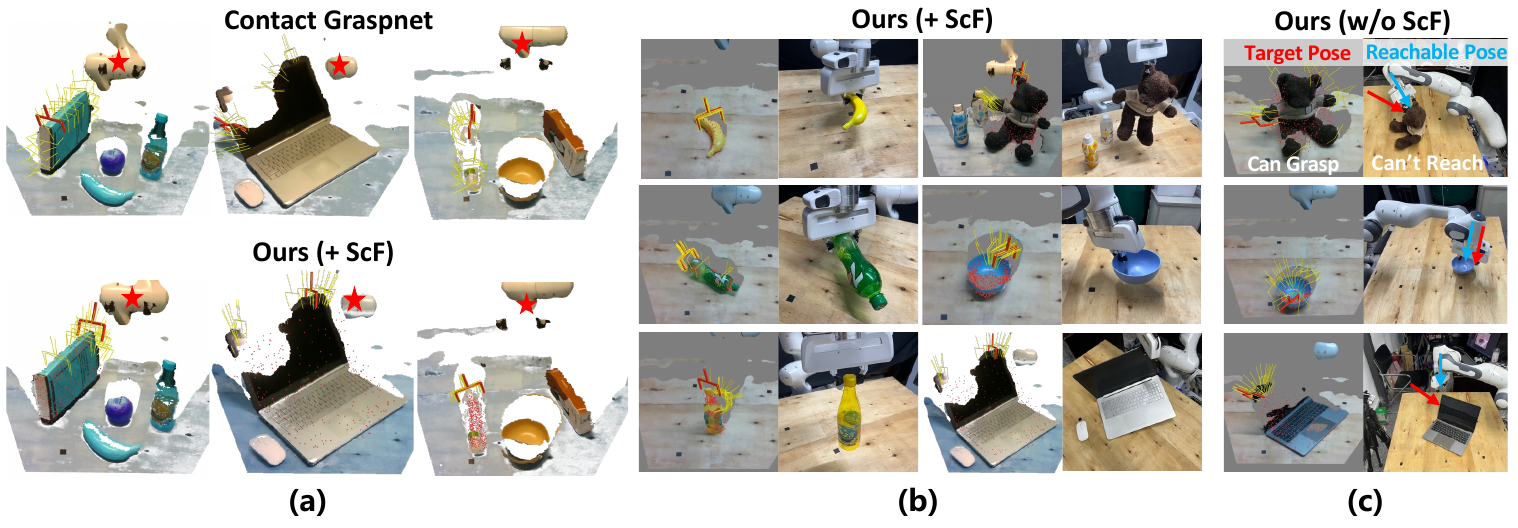}
	\caption{Experiments visualization. (a) shows the predicted grasps of Contact Graspnet \cite{sundermeyer2021contact} and our whole method. Red grasp is the highest grasp score and it will be chosen as the first grasp of our experiment. The location of the red pentagram is the Franka robot. (b) shows the robotic 6-DoF grasp in the real world of our whole method. (c) shows the out-of-workspace failures of ours w/o ScF. The red arrows show the predicted target pose and the blue arrows show the reachable pose. All the grasp proposals of our method are generated based on the original points, and the red completion points are visualized for better understanding.}\label{fig_9}
\end{figure*}

\begin{table*}[!htb]\centering
    \caption{Success Rate of Single Object Grasp Experiments. (Each object has a total of 50 trails with camera opposite to robot.)}
    \label{table_1}
    \begin{tabular}{cccccccccccc}
    \toprule
        {\textbf{Methods}} & {Apple} & {Banana} & {Bear} & {Bowl} & {Mouse} & {Bottles} & {Box} & {Wineglass} & {Laptop} & {Book} & {\textbf{mean(\%)}}\\
    \midrule
    		{6-DoF Graspnet \cite{mousavian20196}} & 48* & 60* & 36* & 42 & 64* & 28 & 50 & 30* & 56* & 46 & 46.0\\
    		{PointNetGPD \cite{liang2019pointnetgpd}} & 62 & 70 & 60* & 56* & 86* & 46 & 66 & 50* & 66* & 58* & 62.0\\
    		{Graspnet-1Billion \cite{fang2020graspnet}} & 66 & 74 & 58 & 56 & 82 & 48 & 68 & 56 & 68 & 60 & 63.6\\
    		{Ma \textit{et al.} \cite{ma2023towards}} & 64 & 70 & 68 & 60 & 88 & 54 & 66 & 52 & 66 & 64 & 65.2\\
    		{Contact Graspnet \cite{sundermeyer2021contact}} & 70 & 76 & 62 & 62 & 92 & 52 & 64 & 56 & 74 & 58 & 66.6\\
    		{Graspness \cite{wang2021graspness}} & 72 & 80 & 64 & 58 & 90 & 62 & 68 & 64 & 68 & 70 & 69.6\\
     \midrule
     	{Beyond Top-Grasps \cite{lundell2019robust}} & 54 & 66 & 44 & 84* & 78 & 64* & 62* & 56 & 60 & 76* & 64.4\\
     	{Trans SC \cite{chen2022improving}} & 78* & 84* & 54 & 88* & 88 & 52 & 84* & 54 & 48 & 78* & 70.8\\
     	{3DSGrasp \cite{mohammadi20233dsgrasp}} & 76* & 84* & 52 & 84* & 90 & 50 & 86* & 56 & 50 & 84* & 71.2\\
     \midrule
        \textbf{Ours} & \textbf{82}* & \textbf{94}* & \textbf{92}* & \textbf{98}* & \textbf{96}* & \textbf{74} & \textbf{94} & \textbf{68} & \textbf{98} & \textbf{94} & \textbf{89.0} \\
    \bottomrule
    \end{tabular}
	\begin{tablenotes}
		\item * means the object is an unknown category object to the corresponding grasp model.
	\end{tablenotes}

\end{table*}

\begin{table*}[!htb]\centering
    \caption{Success Rate of Clutter Removal Experiments. (Remove the objects one by one with 10 trails.)}
    \label{table_2}
    \begin{tabular}{ccccccccc}
    \toprule
        {\textbf{Methods}} & {Set a} & {Set b} & {Set c} & {\textbf{shelter mean}} & {Set 1} & {Set 2} & {Set 3} & {\textbf{close mean}} \\
    \midrule
    		{Contact Graspnet \cite{sundermeyer2021contact}} & 7 & 6 & 6 & 63.3\% & 4 & 4 & 5 & 43.3\%  \\
    		{3DSGrasp \cite{mohammadi20233dsgrasp}} & 5 & 5 & 6 & 53.3\% & 3 & 3 & 5 & 36.7\% \\
    		{Ours} & 7 & 8 & 8 & 76.7\%  & 5 & 3 & 4 & 40.0\% \\
    \bottomrule
    \end{tabular}
\end{table*}

\
\subsubsection{Objects Placed in Clutter} To evaluate whether our method is effective for the clutter situation, which makes the object points incomplete, we place 6-8 random objects on the table in clutter named Set (a-c) to grasp the objects that are sheltered. We also place 6-8 random objects on the table closely cluttered with three sets named Set (1-3) and grasp them one by one to the container box. Each cluttered scene is shown in Fig.11. The camera is opposite the robot in this experiment. We compare our method with the Contact Graspnet\cite{sundermeyer2021contact} and 3DSGrasp\cite{mohammadi20233dsgrasp}. The grasp results are shown in TABLE II. The sheltered clutter leads to the incomplete point of the objects and our methods can remain robust. However, different from the reflective areas, sheltered clutter sometimes with large incomplete as the mask is sometimes half of the target object, which will lead to a large difference in the result of point cloud completion, resulting in a small reduction in the success rate of capturing. Additionally, close clutter requires more consideration of collision avoidance, which is beyond the primary focus of this paper. Like other methods, ours also encounters challenges due to collisions with surrounding objects.

\begin{figure*}[!htb]\centering
	\includegraphics[width=6.8in]{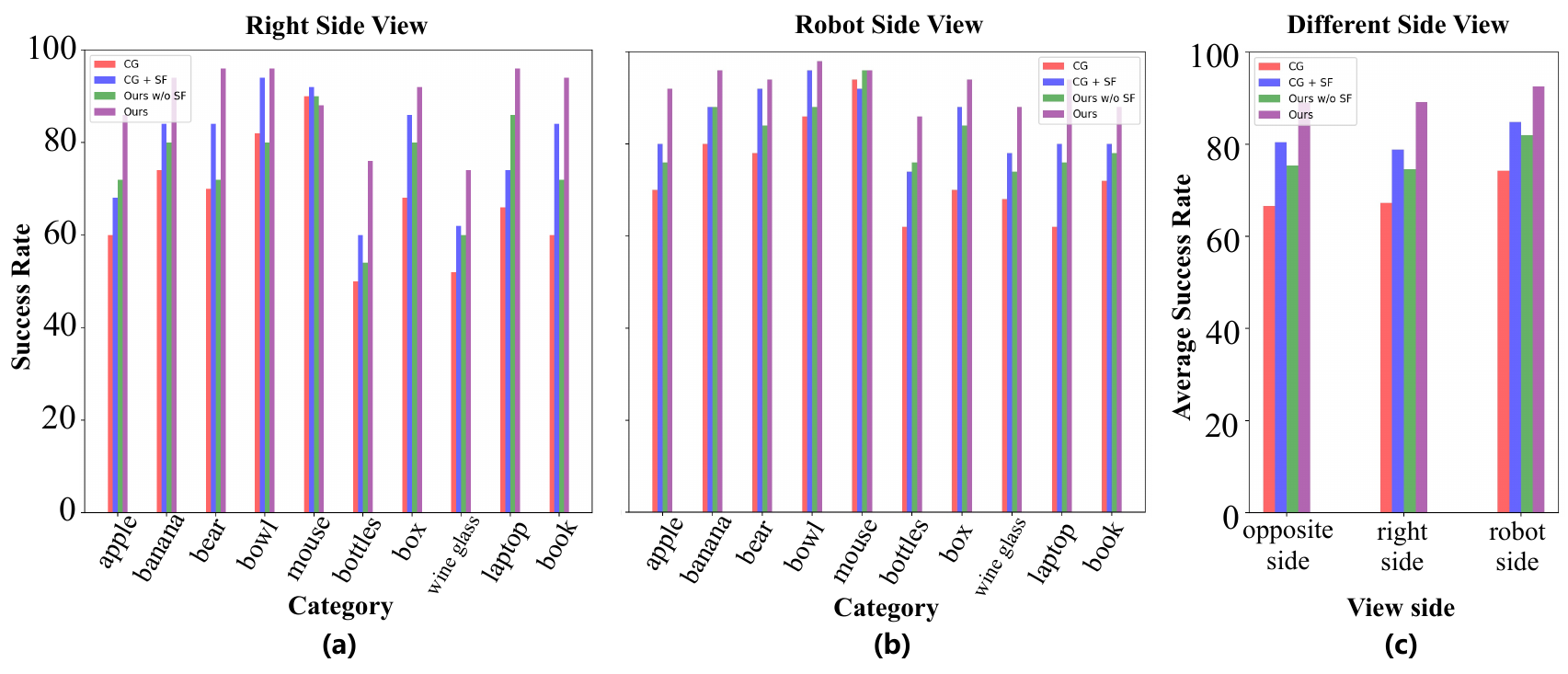}
	\caption{Different view side success rate. (a) and (b) show the right side view and the robot side view success rate of Contract Graspnet \cite{sundermeyer2021contact}, \cite{sundermeyer2021contact} with our ScF, ours w/o ScF and our whole method. (c) shows three camera view's mean success rate of four methods.}\label{fig_11}
\end{figure*}

\begin{figure}[!htb]\centering
	\includegraphics[width=2.3in]{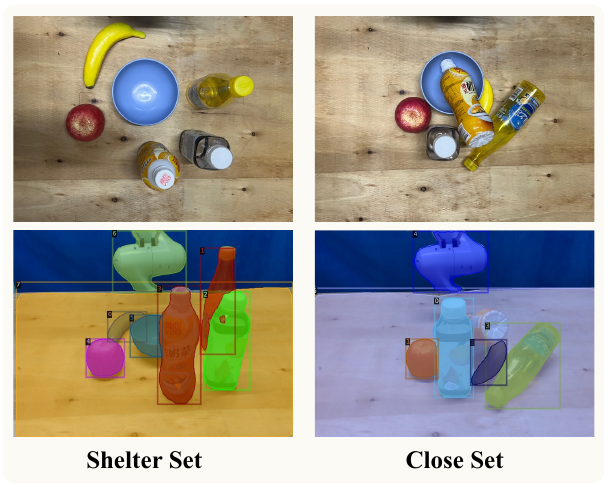}
	\caption{Clutter set experiments. Shelter-set refers to objects placed under cover. Close-set denotes objects positioned closely together without gaps.}\label{fig_10}
\end{figure}

\
\subsubsection{Different View Side of Camera} To further verify the ScF, we also placed the camera on the right side of the object and the robot side of the object. The right side view and the robot side view results are shown in Fig.10. The opposite side view and the right side view get a lower success rate than the robot side view. This is because the grasp poses correspond to the points, and the view approach will influent the grasp pose generation. Thus, the grasp generation in the opposite side view and the right side view is sometimes difficult for the robot to execute. However, our method keeps about a 90\% success rate in all the camera views.

We record the failure numbers out of the robot workspace in Table I experiments about our method and the baseline method in Fig.12. All of the objects are placed in the robot workspace as Fig.8 shows unless the robot grasps them in the wrong way. Our ScF almost removes all the inconvenient grasps. In the first stage of our method, the network predicts grasp poses and corresponding confidence scores based on the geometric shape of the object. In the second stage, ScF adjusts the confidence scores by considering the spatial relationship between the robot arm and the grasp pose. This approach takes both the object's geometry and the robot's motion planning into account. As a result, the scores of less feasible grasps are reduced, allowing for the selection of grasps that are better suited for the robot arm to execute.

However, there are instances where a pose may be easier to grasp but the object's geometry makes it difficult. For instance, the bear is face down and its feet are back to the robot, which is inconvenient for the robot to grasp its back. Therefore, the high-score grasps are back to the robot even after the filter. In such cases, the scores for these easier-to-execute grasps are relatively low, and the high-scoring grasps predicted by the network remain higher after filtering, which causes the ScF inefficient.

\subsection*{C. Ablation Study}
\setcounter{subsubsection}{0}
\
\subsubsection{Effect of Point Completion Numbers} Since we use point completion to provide complete object shape information, we verify its influence on the grasping of the completion number. We contrast 512, 1024, 2048 and 4096 to separately concatenate with the original points. We train Contact Graspnet as the baseline method. The results of the grasp training loss are shown in Fig.13. Notably, 512 points are sampled with FPS from $\mathbf P_{c}$, 2048 and 4096 points are sampled with FPS from 16384 points of refine prediction $\mathbf P_{ref}$ based on the coarse points follow the PCN\cite{yuan2018pcn}. Our methods get lower loss results than Contact Graspnet (2048\_CG). It can be seen that the 512 points get higher losses than 1024 points or more, suggesting that the coarse prediction of 1024 points from PCN is enough in our method. Unlike traditional point completion, which aims to use as many points as possible to provide detailed object information, we use point completion to offer shape information crucial for grasping. Therefore, the primary goal of point completion in our context is to provide accurate shape information, while detailed shapes are not essential for grasping tasks. Using sparse points leads to significant deviations in object shape. However, once the point number reaches a certain threshold, further increasing the number of points does not significantly enhance the representation of the object's major shape. Consequently, 1024 points are adequate for effective grasping and inference.

\begin{figure}[!htb]\centering
	\includegraphics[width=3.2in]{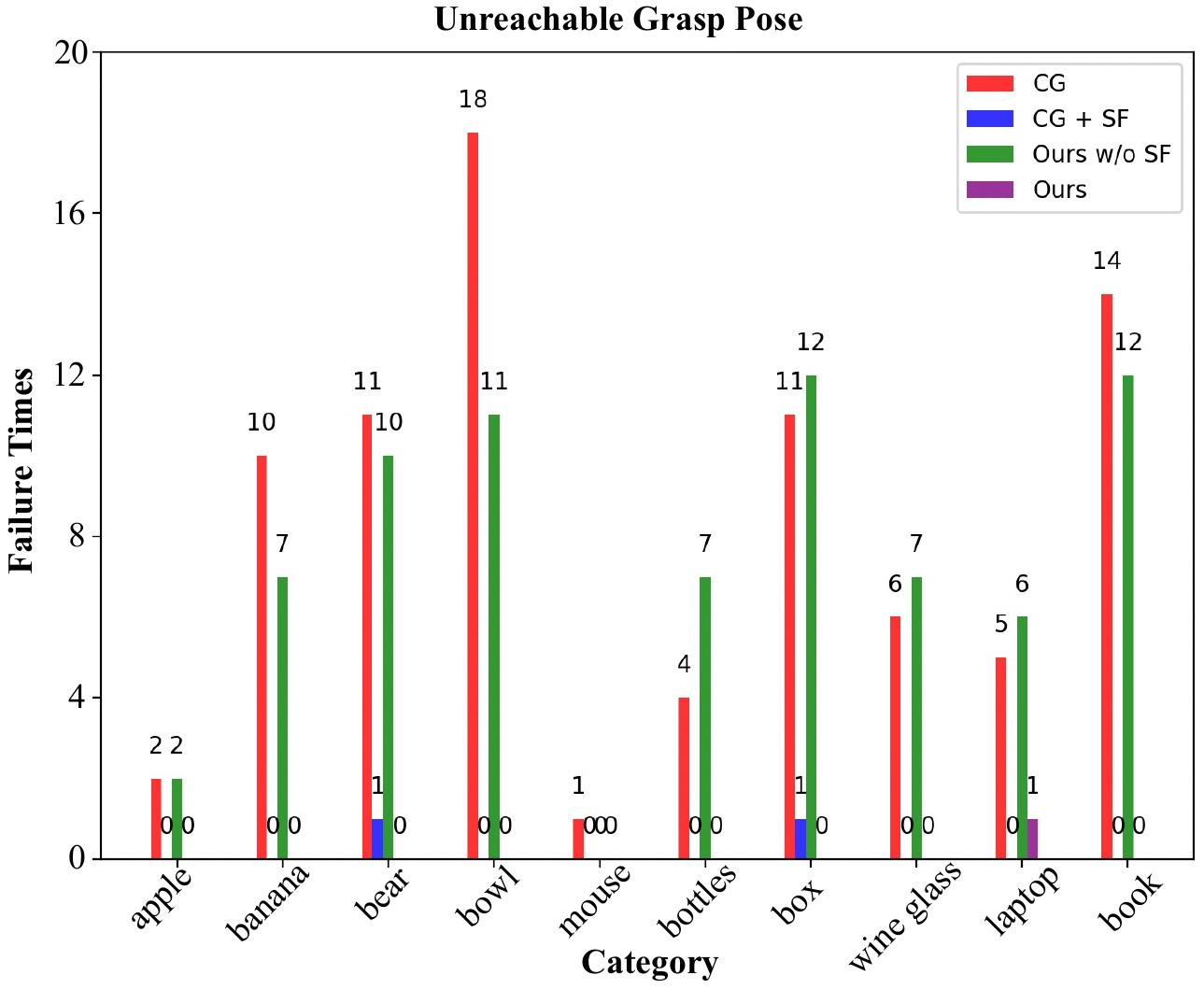}
	\caption{Failure numbers out of workspace. For each object, red is the Contact Graspnet's (CG) result, blue is the CG + ScF, green is our vanilla network (Ours w/o ScF), and purple is our whole method.}\label{fig_12}
\end{figure}

\begin{figure}[!htb]\centering
	\includegraphics[width=3.2in]{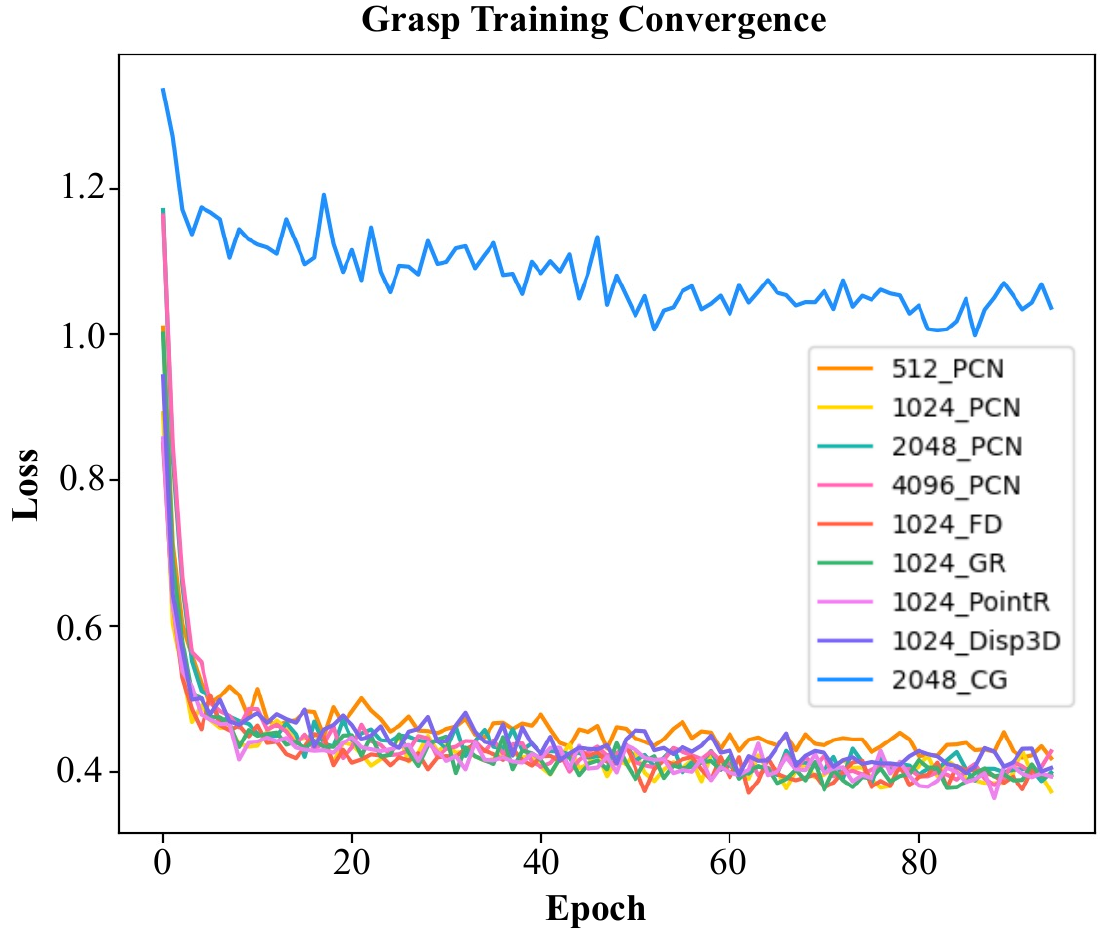}
	\caption{Training loss of different methods. 512\_PCN, 1024\_PCN, 2048\_PCN and 4096\_PCN are different numbers of points of PCN\cite{yuan2018pcn}. 1024\_FD, 1024\_GR, 1024\_PR and 1024\_D3D are 1024 points of FoldingNet\cite{yang2018foldingnet}, GRNet\cite{xie2020grnet}, PointR\cite{yu2021pointr} and Disp3D\cite{wang2022learning}. 2048\_CG is method of Contact Graspnet\cite{sundermeyer2021contact}.}\label{fig_13}
\end{figure}

\begin{table*}[!htb]\centering
    \caption{Ablation Results of Single Object Grasp Experiments. (Each object has a total of 50 trails with camera opposite to robot)}
    \label{tab:univ-compa}
    \begin{tabular}{cccccccccccc}
    \toprule
        {\textbf{Methods}} & {Apple} & {Banana} & {Bear} & {Bowl} & {Mouse} & {Bottles} & {Box} & {Wineglass} & {Laptop} & {Book} & {\textbf{mean(\%)}}\\
    \midrule
        {Baseline} & 66.0 & 82.0 & 60.0 & 64.0 & 86.0 & 48.0 & 66.0 & 50.0 & 68.0 & 60.0 & 65.0\\
    \midrule
    		{PC r/b FDNet\cite{yang2018foldingnet}} & 84.0* & 88.0* & 90.0* & 92.0* & 94.0* & 76.0 & 90.0 & 62.0 & 92.0 & 96.0 & 86.4\\
    		{PC r/b GRNet\cite{xie2020grnet}} & 88.0* & 92.0* & 90.0* & 94.0* & 90.0* & 72.0 & 96.0 & 70.0 & 88.0 & 92.0 & 87.2\\
    		{PC r/b PointR\cite{yu2021pointr}} & 90.0* & 96.0* & 86.0* & 98.0* & 94.0* & 78.0 & 96.0 & 70.0 & 94.0 & 92.0 & \textbf{89.4}\\
    		{PC r/b Disp3D\cite{wang2022learning}} & 88.0* & 86.0* & 92.0* & 96.0* & 98.0* & 70.0 & 92.0 & 72.0 & 94.0 & 90.0 & 87.8\\
     \midrule
     		{Ours b/o completion} & 56.0* & 90.0* & 80.0* & 94.0* & 94.0* & 66.0 & 80.0 & 64.0 & 80.0 & 78.0 & 78.2\\
     		{Contact Graspnet \cite{sundermeyer2021contact} \textbf{+ ScF}} & 76.0 & 84.0 & 90.0 & 82.0 & 90.0 & 66.0 & 86.0 & 64.0 & 90.0 & 76.0 & 80.4\\
     	\textbf{Ours w/o PCF} & 46.0* & 90.0* & 42.0* & 96.0* & 96.0* & 72.0 & 84.0 & 62.0 & 68.0 & 84.0 & 74.0 \\
        \textbf{Ours w/o ScF} & 78.0* & 78.0* & 66.0* & 80.0* & 94.0* & 62.0 & 74.0 & 62.0 & 88.0 & 72.0 & 75.4 \\
        \textbf{Ours} & 82.0* & 94.0* & 92.0* & 98.0* & 96.0* & 74.0 & 94.0 & 68.0 & 98.0 & 94.0 & \textbf{89.0} \\
    \bottomrule
    \end{tabular}
	\begin{tablenotes}
		\item * means the object is an unknown object to the corresponding grasp model.
	\end{tablenotes}

\end{table*}

\
\subsubsection{Effect of Point Completion Accuracy} In point completion, each point's CD loss would accumulate a large bias in whole points. Therefore, we evaluate the point completion accuracy with four more point completion works FoldingNet\cite{yang2018foldingnet}, GRNet\cite{xie2020grnet}, PointR\cite{yu2021pointr} and Disp3D\cite{wang2022learning}, all the method are trained as the implementation of point completion details. We use 1024 points as the completion points and train our grasp network. The results of training loss are shown in Fig.13. We also implement the grasp experiments in the opposite-side view of these methods and the results are shown in TABLE III. As all the real-world objects are not trained in the dataset, and the input points are extremely incomplete, the outcomes of different point completion methods approximate the true shapes of the objects, yielding only similar shape information. This is the reason why the training loss in Fig.13 and success rate in TABLE III change little.

\
\subsubsection{Effect of Different Modules}

To verify different modules in our proposed framework, we execute the same real robot experiments as TABLE I, and the results are shown in TABLE III. The results of Ours w/o PCF directly input the original points are 15\% lower than the Ours, which demonstrates the point completion results provide better object information for grasping. Furthermore, the success rate of grasping performed directly with completion points (Ours b/o completion) is about 10.8\% lower than that of features (Ours). This experiment evaluates that directly utilizing the completion points becomes ineffective when the completion results lack precision. Therefore, objects with large volumes, such as the box, laptop, bear and book, lead to the grasp proposals that are sometimes far away from the actual objects. This is because the points will be distributed among both the target object points and incorrect completion object points. The apple has a non-negligible bias in grasp position as the grasp pose will collide with the apple. Only the banana, bowl, wineglass and mouse have thin geometry that can be easily grasped with bias.

\begin{table}[!htb]\centering
    \caption{Inference Efficiency of Grasp Models.}
    \label{tab:univ-compa}
    \begin{tabular}{ccccccccc}
    \toprule
        {methods} & {params} & {Time on SC} & {Time on RC} \\
    \midrule
    		{Contact Graspnet \cite{sundermeyer2021contact}} & 27.9M &  0.39 $\pm$ 0.02s & 1.7 $\pm$ 0.05s \\
    		{3DSGrasp \cite{mohammadi20233dsgrasp}} & 505.7M & 1.02 $\pm$ 0.05s & 3.5 $\pm$ 0.14s \\
    		{Ours} & 24.3M + 33.7M & 0.14 $\pm$ 0.02s & 1.1 $\pm$ 0.07s \\
    \bottomrule
    \end{tabular}
\end{table}

The experiments shown in TABLE III evaluate the effectiveness of our ScF, which selects reachable grasps for the robot. Fig.9 (c) shows the out-of-workspace of ours w/o ScF. It can be seen that the grasp pose is opposite to the robot itself, which is hard for a robot to reach, despite some poses being good enough for a robot to execute. Conversely, our score filter will always choose the pose convenient for a robot to grasp in Fig.9 (b). What's more, as we apply our ScF to the Contact Graspnet\cite{sundermeyer2021contact} with the same grasp representation, it improves the success rate by 14\% than TABLE I, which proves that our ScF can be a paradigm for 6-DoF grasp network to real robot execution.

\subsection*{D. Network Efficiency}

We show the model size and inference time in TABLE IV, which is the same in both isolated and cluttered situations. Although our method has a point completion model and a grasp model, the reason why we are faster than Contact Graspnet\cite{sundermeyer2021contact} is our PCF-Layer does not sample points with FPS, which takes about $0.45s$ due to its extensive looping and iterations. But Contact Graspnet directly use PointNet++ to sample 2048 points from 20000 points wastes a lot of time. What's more, we do some experiments to equally verify the computation speed. We translate the Contact Graspnet code from Tensorflow to Pytorch (CGtorch), and we also feed 1024 points into Contact Graspnet (CG1024). The inference speed is $1.75 \pm 0.05s$ of CGtorch and $1.67 \pm 0.04s$ of CG1024, which are still slower than our method. 3DSGrasp\cite{mohammadi20233dsgrasp} has twice major KNN iteration process and larger transformer models, leading it more time cosuming.

\section{Discussion and Conclusion}

In this study, inspired by human grasping, we propose a novel 6-DoF grasp framework based on the point completion feature, which converts the coarse point completion results as a shape feature into hidden space to train the grasp network. The results demonstrated that our framework improves grasp accuracy and can be used to grasp unknown objects. Furthermore, a score filter is integrated into our framework, which determines the optimal grasp pose for real-world execution, thereby eliminating the need for applying network output to the physical robot. Extensive experiments demonstrate that the proposed framework achieves state-of-the-art performance even though the objects are challenging for a parallel gripper. We think that integrating the RGB feature into the point cloud to augment the contextual information would be beneficial for enhancing grasping performance, and we plan to investigate this in our future work.

Overall, our method offers partial evidence that supplying the network with more comprehensive information enhances the reliability of its outputs. This finding is particularly enlightening for related fields, suggesting that the conversion of valuable yet complex information into accessible hidden features represents a strategic approach.

\bibliographystyle{IEEEtranTIE}
\bibliography{IEEEabrv, main}

\end{document}